\documentclass[10pt,twocolumn,letterpaper]{article}

\usepackage[cvprfinal]{cvpr}      %
\usepackage{cvpr}              %

\usepackage{graphicx}
\usepackage{amsmath}
\usepackage{amssymb}
\usepackage{booktabs}
\usepackage{svg}
\usepackage{tabularx} %
\usepackage{float}

\usepackage[pagebackref,breaklinks,colorlinks]{hyperref}

\usepackage[font=small]{caption}

\usepackage[capitalize]{cleveref}
\crefname{section}{Sec.}{Secs.}
\Crefname{section}{Section}{Sections}
\Crefname{table}{Table}{Tables}
\crefname{table}{Tab.}{Tabs.}

\def\cvprPaperID{1234} %
\def\confName{CVPR}
\def\confYear{2022}

\begin{document}

\title{SpotNet: An Image Centric, Lidar Anchored Approach To Long Range Perception}

\author{Louis Foucard, Samar Khanna, Yi Shi, Chi-Kuei Liu, Quinn Z Shen, Thuyen Ngo, Zi-Xiang Xia \\
Aurora Innovation 
{\tt\small flouis@aurora.tech}
}

\maketitle

\begin{abstract}
In this paper, we propose SpotNet: a fast, single stage, image-centric but LiDAR anchored approach for long range 3D object detection. 
We demonstrate that our approach to LiDAR/image sensor fusion, combined with the joint learning of 2D and 3D detection tasks, can lead to accurate 3D object detection with very sparse LiDAR support. Unlike more recent bird's-eye-view (BEV) sensor-fusion methods which scale with range $r$ as $\mathcal{O}(r^2)$, SpotNet scales as $\mathcal{O}(1)$ with range.
We argue that such an architecture is ideally suited to leverage each sensor's strength, i.e. semantic understanding from images and accurate range finding from LiDAR data. 
Finally we show that anchoring detections on LiDAR points removes the need to regress distances, and so the architecture is able to transfer from 2MP to 8MP resolution images without re-training.
\end{abstract}

\section{Introduction}
\label{sec:intro}

Ensuring accurate 3D detections over extended distances is pivotal for the safe operation of autonomous trucks. Considering the potential load of up to 80,000 lbs and challenging road conditions like wet or icy surfaces, trucks might require a stopping distance of nearly 230m \cite{fmcsa}. Notably, truck operators frequently make decisions based on what they perceive far beyond these ranges, such as deciding to change lanes away from disabled shoulder vehicles as required by the Move Over law \cite{move_over_law}. While such preemptive maneuvers might not always be safety-critical, these early decisions diminish the likelihood of urgent, critical situations. Therefore, in addition to maintaining an ability to detect at the fully loaded stopping distance, to ensure naturalistic driving, autonomous vehicles need to be able to detect and react to objects far beyond the stopping distance. However, these distances surpass the sensory range of a majority of LiDAR systems, and are far beyond the ranges of ~100m considered in most public 3D detection benchmarks. Recent developments in specialized long-range lidars such as the Frequency Modulated Continuous Wave (FMCW) systems \cite{FMCW}\cite{Aurora}\cite{Aurora2} can achieve maximum sensing range of more than 400m, but the sparsity of the resulting lidar point clouds at these distances makes the 3D detection task particularly challenging.

Perception systems in autonomous driving have made immense progress over the last few years, largely thanks to the advent of deep neural networks and larger datasets. This progress is also due to the introduction of new sensor fusion approaches that leverage the high spatial resolution of cameras and the precise 3D information of lidar sensors. A large majority of the proposed sensor fusion methods adopt a lidar-centric approach, which rely on the presence of dense lidar points on the objects in order to produce detections. Although lidar sensors have improved considerably, they are still orders of magnitude away from the spatial or temporal resolution that cameras achieve at a fraction of the cost, and can be largely blind to some non-lambertian and/or low albedo surfaces (e.g. black car). This strong reliance on lidar introduces safety-critical failure modes in situation that require the detection of objects at long ranges with very few lidar returns. Lidar-centric approaches might also not be best suited to detect in 3D the numerous other classes of objects needed for scalable autonomous driving, such as signs, traffic cones,  traffic lights or other objects with small sizes and strong textural and visual features.

A majority of lidar-centric sensor fusion approaches adopt a cell based representation of the data, where the lidar point cloud is projected and discretized into a 2D grid of fixed dimensions \cite{Liang_2018_ECCV_cont_fuse, pmlr-v87-yang18b_hdnet, pmlr-v87-casas18a_intentnet} or in 3D voxels \cite{zhou2017voxelnet}. This representation has several advantages such as range invariance, and a natural framework in which to bring multiple lidar sweeps and temporal dependence. However, this comes at the cost of a lower spatial resolution as multiple lidar points are aggregated into the same spatial location, and heavy computational costs that scale as $\mathcal{O}(r^2)$ with range $r$.

In order to maintain the lidar spatial resolution as much as possible, as well as scale as $\mathcal{O}(1)$ with range $r$, a  range view (RV) representation of the lidar data was adopted in \cite{meyer2019lasernet,meyer2019lasernet++,chen2016mv3d}, where a dense lidar image is created using cylindrical coordinates. These methods are more computationally efficient, and retain all the lidar spatial resolution. Finally, point based methods \cite{qi2016pointnet, qi2017pointnet++} process lidar data directly in their original 3D point cloud space, avoiding both the sparsity and loss of information from discretization, but again at large computational costs that does not scale well with range.

In all above mentioned approaches, both the image feature fusion and 3D region proposals are strongly dependent on the presence of lidar points on the objects of interest. This lidar-centric view results in sensor fusion architectures that discard the large majority of image data. Recently, more image-centric sensor fusion approaches have been proposed \cite{Xu_2018_CVPR_MLF, wang2018pseudolidar} that use monocular depth estimation in order to generate a pseudo lidar point cloud from images, which can be processed by any aforementioned lidar methods, while \cite{Liang_2019_CVPR_mmf} performs depth completion on the original lidar point cloud in order to perform dense image feature fusion. However, these methods rely on the computationally costly and ill-posed monocular depth estimation for the entire scene - when only depth on the detected objects should be necessary. Early errors in depth estimation can severely hamper image/lidar feature fusion. 

A lesser utilized, image-centric sensor fusion approach consists in projecting lidar data into the image, and processing the data as an RGB-D image. This sensor fusion approach is computationally efficient and more robust to lidar failure, as shown by \cite{Pfeuffer2019RobustSS} for semantic segmentation in adverse weather condition. 

To address the issues mentioned above and fully leverage the respective strength of camera and lidar data, we propose the following method: (1) in order to keep inference time $\mathcal{O}(1)$ with respect to range and preserve all sensor data, we choose a range view RGB-D sensor fusion scheme with a sparse depth raster channel. (2) to improve the use and quality of the extracted image features, we formulate the detection problem as a multi-modality 2D/3D detection task and supervise all losses in image space. (3) to retain the 3D accuracy of lidar data, we anchor both 2D and 3D detection on lidar points. Our contributions are as follows:
\begin{enumerate}
    \item We present a range-view approach that scales efficiently with range and adapts sensor fusion from a sparse image, dense lidar representation to a dense image, sparse lidar RGB-D representation that retains all sensor data.
    \item We show how multi-modality labeling and joint 2D and 3D detections multitask supervision improves long range detection performance.
    \item Finally we argue that our method is uniquely suited to take advantage of high resolution imagery, and show that the approach is able to transfer well from 2MP to 8MP images, improving long range performance beyond label ranges without retraining. 
\end{enumerate}

\section{Related Work}
\label{sec:related_work}

Object detection has rapidly developed in the last few years due to advances in deep learning, sensors, compute platforms, and availability of datasets. While 2D detection has matured the most, 3D detection, especially at longer ranges, remains an open area and is being more actively explored today.

\subsection{Image-centric 3D detection}

3D detection using only cameras is inherently difficult due to poor depth estimation from images, resulting in poor 3D localization. Stereo approaches estimate depth based on disparity between corresponding points in a pair of stereo cameras, but often at significant computation and complexity cost \cite{Li_2019_CVPR, qin2019triangulation}. Failure modes tend to be low texture areas where point matching and disparity estimation are difficult. Monocular 3D depth estimation aim to generate a pseudo lidar point cloud from images, which can be processed by 3D detection methods \cite{Xu_2018_CVPR_MLF, wang2018pseudolidar}. But this is even more challenging due to the ill posed nature of the problem, and lack of any direct depth measurements. Nonetheless, significant progress has been made in this area. Some methods are based on direct 3D proposal generation. Mono3D \cite{he2019mono3d} directly generates such proposals and scores them using various appearance, geometry features as well as a prior belief that the object is on the ground plane. Others leverage inherent geometric constraints that arise from relationship between 2D and 3D boxes: Deep3DBox \cite{mousavian20173d} uses established 2D detection methods to infer a 2D box, and then seeks to find a 3D box that fits tightly along at least one of the sides of the 2D box. More recently, mono depth methods are exploiting keypoints and shape in order to jointly reason over appearance and geometry \cite{bertoni2019monoloco, qin2018monogrnet}. MonoGRNet used 4 subnetworks to reason jointly on 2D detection, instance depth, 3D location, and local corner regression. Finally, \cite{zhou2019objects} models object as centers of their bounding box, and directly regresses 3D extents and pose in camera frame as an additional parameterization to obtain 3D detections from as single image.
\subsection{Lidar-centric 3D Detection}
Lidar offers direct 3D sensing, but at a much lower resolution and density than cameras. It also does not provide as rich visual information such as color and texture. This makes detection of smaller and/or farther objects difficult as such objects may have only a few lidar points on them. Multiple approaches have accomplished 3D detection suitable for urban scenes from lidar point clouds \cite{zhou2017voxelnet,qi2016pointnet,lang2018pointpillars, shi2018pointrcnn}, as well as combined detection, tracking, and prediction from sequences of point clouds \cite{Luo_2018_CVPR_FaF, pmlr-v87-casas18a_intentnet, Liang_2020_CVPR_pnpnet}. Voxelization based methods \cite{zhou2017voxelnet} discretize space into 3D voxels, which preserves 3D shape information but results in many empty voxels due to sparse point clouds. Other methods \cite{qi2016pointnet,qi2017pointnet++} avoid voxelization and operate on sets of points directly, learning and inferring features for each point directly.  The operations can be on the discrete set of points \cite{qi2016pointnet, qi2017pointnet++, shi2018pointrcnn} or specially designed for continuous space such as parametric continuous convolution \cite{Wang_2018_CVPR_deep_cont_conv}. Other methods project the lidar points into a 2D grid based feature map that is then analyzed with 2D convolutions. The target view can be range view \cite{meyer2019lasernet} or birds eye view (BEV) \cite{lang2018pointpillars, atglasernet}. While these methods have achieved quite high performance for vehicles, they face more challenges for smaller classes like pedestrians and cyclists, especially at higher range. Moreover, the use of a bird's eye view grid means that the inference time will scale with the square of the distance, making such approaches impractical for real-time long range 3D object detection. 
\subsection{Multimodal 3D Detection}
Multimodal 3D detection aims to take fuller advantage of cameras and lidars. The design space is larger due to the questions of when and how to fuse, as well as choice of view (BEV vs. range view), and single stage vs. two stage. The fundamental tradeoff has been point density and information content (camera) vs. direct 3D measurement per point (lidar). 
PointPainting \cite{vora2019pointpainting} proposed a sequential approach which appends lidar points with semantic labels from a 2D camera based segmentation task. This has shown significant improvement over several lidar only methods, however suffers from parallax between lidar and camera sensors, as well as point mixing from object transparency.
MV3D \cite{chen2016mv3d} and AVOD \cite{ku2017avod} have both applied a 2 stage, 2 stream mid level fusion approach  to extract features, propose regions, and perform detection. However, the ROI feature fusion and region proposals happen at higher level feature maps, and the 2 stream 2 stage architecture can be computationally demanding. Continuous Fusion \cite{Liang_2018_ECCV_cont_fuse} addresses these bottlenecks using continuous convolution to fuse multiple convolutional feature maps from both sensor streams in BEV space. While improving detection performance in general, it still is subject to the fundamental problem of sparse lidar points especially at distance. MMF \cite{Liang_2019_CVPR_mmf} proposed an efficient multitask single stage multimodal architecture that employs two backbones with dense fusion. It added a dense depth task to help address the lidar sparsity bottleneck. More recently, \cite{liang2022bevfusion} disentangles the lidar and camera feature extractor, such that the presence of camera features does not depend on lidar data, and fuses both into a BEV feature map. This enables the model to be robust to lidar malfunctions during severe weather. However, due to the use of BEV for feature aggregation, this approach is not able to scale efficiently at longer ranges for real time applications.

\section{Proposed Method}

\label{sec:method}
We describe here in detail our proposed approach, from sensor data fusion, to losses, and output decoding. 
\begin{figure*}[htbp]
  \centering
  \includegraphics[width=\linewidth]{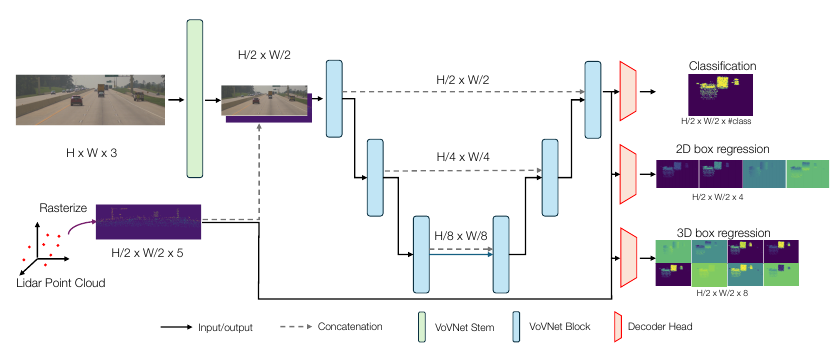}
  \caption{Overall architecture diagram. The figure above shows the input RGB-D data as well as the output maps for each of the category, 2D regression, and 3D regression head. The lidar data is rasterized and used both in early and late sensor fusion.}
\end{figure*}
\subsection{Sensor Fusion}
The camera and lidar data is fused using a RGB-Depth approach, by projecting the lidar point cloud into the image at half resolution ($160\times790$ for 2MP, $320\times1580$ for 8MP images) and forming sparse depth rasters. We use z-buffering in order to mask out points that become occluded when transforming from the lidar to camera frame (or from double returns). In practice, the resolution of the image into which we project the lidar data is large enough that only a small fraction of the points ($< 1\%$) end up being suppressed by z-buffering, especially when running on 8MP images. The sparse depth raster contains two channels: the Euclidean distance from the camera frame to the points, and a binary sentinel channel to indicate which pixels contain a valid lidar return. It is then resized and concatenated to the RGB channels of the image tensor, forming a five channel input tensor (see Figure 1). As detailed in the network section below, the lidar data is fused and injected in the network at various stages, including just before the final decoding head. The depth raster is then resized to the appropriate resolution using closest neighbor sampling in order to be concatenated with the image or feature channels of various resolutions.

\subsection{Network Architecture}
The RGB-D input tensor at the original image resolution (2MP or 8MP) is first fed into a stem network composed of 2 fully convolutional layers with 32 and 64 dimensions with kernel sizes of $7\times7$ and $3\times3$ respectively. The first layer has a step size of 2, bringing down the feature resolution to half the original resolution.

This output of the stem is then concatenated with the depth raster at half resolution, then fed into a VoVNetV2 \cite{lee2020centermask} feature extractor with three stages. The first three each stages apply a $2\times$ downsampling, while the last three up-sample the feature map back to half resolution. At each of the last 3 up-sampling stages, the depth raster is resized and concatenated with the feature map before being fed into the next upsampling stage.
Finally, the depth raster is concatenated one last time with the output feature map of the last stage of the VoVNetV2 trunk and fed into the decoding heads. Each decoding head consists in a $1\times1$ convolution layer, with output activation and dimension detailed in the section below.

\begin{figure*}[htbp]
  \centering
  \includegraphics{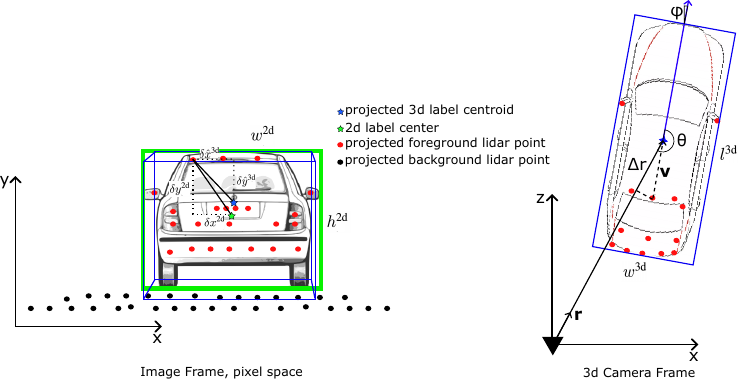}
  \caption{2D and 3D target encoding. On the left, the 2D labels and projected 3D labels are shown in the image frame in pixel space, along with projected foreground (red) lidar points and background (black) lidar points. On the right, the 3D label is shows in top down view in 3D camera frame, along with foreground lidar points.}
\end{figure*}
\subsection{Target Encoding and Losses}

The feature extraction stage of the network produces an output at half the resolution $H/2 \times W/2$ of the original input. This output is then concatenated with the depth raster and fed to three distinct decoding branches: the class head, the 2D bounding box head, and the 3D bounding box head. The network is optimized to predict, for each pixel in the output grid, a class probability and to parameterize bounding boxes in both 2D image frame and 3D camera frame. The supervision of these predictions is conducted in a sparse manner, applying losses solely on pixels that correspond to valid lidar point projections. This is implemented by masking the loss function in the $H/2 \times W/2$ output grid, utilizing the sentinel channel from the depth raster at the corresponding resolution to identify relevant pixels.

Consider a set $\mathcal{P}$ of $N$ lidar points and a set $\mathcal{L}$ of $M$ linked 2D and 3D labels, with known correspondences between lidar points and labels. For a given object $k \in \mathcal{L}$, and a lidar point $i \in \mathcal{P}$ that belong to the object, the 2D bounding box parameters are defined as follows. The displacement in pixels from the projected point's location in the image to the center of the object's 2D bounding box is given by $\delta x_{ik}^{\text{2D}}$ and $\delta y_{ik}^{\text{2D}}$:
\begin{equation}
\begin{aligned}
\delta x_{ik}^{\text{2D}} &= x^{\text{2D}}_{k} - \hat{x}_{i} \\
\delta y_{ik}^{\text{2D}} &= y^{\text{2D}}_{k} - \hat{y}_{i}
\end{aligned}
\end{equation}
where $(x^{\text{2D}}_{k}, y^{\text{2D}}_{k})$ denote the center coordinates of the 2D bounding box for object $k$, and $(\hat{x}_{i}, \hat{y}_{i}) = P(x_i, y_i, z_i)$ are the pixel coordinates of the projected lidar point $(x_i, y_i, z_i)$, with $P$ the projective transformation from 3D camera frame to image frame. In the remainder of the article, we will use $\hat{}$ to denote values obtained through projective transformations. Additionally, the width $w^{\text{2D}}_{k}$ and height $h^{\text{2D}}_{k}$ of the 2D bounding box are also part of the parameterization.

The 3D bounding box parameters account for the discrepancy between the 2D box center and the projected centroid of the 3D label due to the non-orthographic projection. For each lidar point $i$ associated with a 3D label $k$, the offsets $\delta \hat{x}_{ik}^{\text{3D}}$ and $\delta \hat{y}_{ik}^{\text{3D}}$ represent the pixel distance from the point's location to the projected centroid of the 3D box:
\begin{equation}
\begin{aligned}
\delta \hat{x}_{ik}^{\text{3D}} &= \hat{x}^{\text{3D}}_{k} - \hat{x}_{i} \\
\delta \hat{y}_{ik}^{\text{3D}} &= \hat{y}^{\text{3D}}_{k} - \hat{y}_{i}
\end{aligned}
\end{equation}
where $\hat{x}^{\text{3D}}_{k}$ and $\hat{x}^{\text{3D}}_{k}$ are the pixel coordinates of the projected 3D label centroid for object $k$. In addition to the above projection displacement, the distance delta from point $i$ to the 3D centroid of object $k$ is parameterized as the dot product 
\begin{equation}
\begin{aligned}
\delta d_{ik} = \mathbf{r}_{ik} \cdot \mathbf{v}_{ik}.
\end{aligned}
\end{equation}
Here, $\mathbf{r}_{ik}$ denotes the unit vector along the camera ray pointing towards the centroid of object $k$, and $\mathbf{v}_{ik}$  is the displacement vector from point i to the centroid of object k in 3D camera frame.
The heading of the object $\phi_k$ is parameterized with respect to the bearing to the object, as $\cos{\theta_k}, \sin{\theta_k}$ with $\theta_k = \phi_k - \alpha_k$, with $\phi_k$ the heading in camera frame and $\alpha_k$ the bearing to the object $k$. This ensures consistency across different viewpoints. Finally, the object's 3D extents are parameterized directly as $w^{\text{3D}}, l^{\text{3D}}, h^{\text{3D}}$ ((see Figure 2).

We use focal loss on the predicted category probability:
\begin{equation}
\mathcal{L}_{\text{class}} = -\frac{1}{N} \sum_{i}^{N} \alpha_i (1 - p_i)^{\gamma} \log(p_i)
\end{equation}
where $N$ denotes the number of pixels associated with valid lidar point projections, $p_i$ represents the predicted probability for the true class at pixel $i$, and $(\alpha,\gamma)$ are focal loss parameters \cite{Retinanet}. 
To train the network to handle uncertainty in the regressed position and extent parameters of the 2D/3D bounding boxes, the network predicts both the mean and the diversity of a Laplacian distribution for each parameter. The supervision for these predictions involves minimizing the negative log-likelihood of the Laplacian distribution, resulting in the following losses for the 2D/3D centers and extents:
\begin{equation}
\begin{aligned}
\mathcal{L}_{\text{2D}} &=& \frac{1}{N} \sum_{i=1}^{N} \left( \frac{\|\delta x^{\text{2D}}_{i} - x_{i}^{\text{2D}*} \|_{1}}{b_{x_i^{\text{2D}}}} + \frac{\|\delta y^{\text{2D}}_{i} - y_{i}^{\text{2D}*} \|_1}{b_{y_i^{\text{2D}}}}  \right. \\
& + & \left. \frac{\| w^{\text{2D}}_{i} - w_{i}^{\text{2D}*} \|_1}{b_{w_i^{\text{2D}}}} + \frac{\| h^{\text{2D}}_{i} - h_{i}^{\text{2D}*} \|_1}{b_{h_i^{\text{2D}}}} \right. \\ 
& + & \left.  \log(b_{x_i^{\text{2D}}}b_{y_i^{\text{2D}}}) + \log(b_{w_i^{\text{2D}}}b_{h_i^{\text{2D}}})\right) 
\end{aligned}
\end{equation}
where $^*$ denotes the model's predictions. The losses $\mathcal{L}_{\text{3D}}$ for the 3D position and extent parameters are computed in an analogous way, while the orientation estimates are supervised using L1 loss.
\begin{equation}
\begin{aligned}
\mathcal{L}_{\text{3D}} &=& \frac{1}{N} \sum_{i=1}^{N} \left( \frac{\|\delta \hat{x}^{\text{3D}}_{i} - x_{i}^{\text{3D}*} \|_{1}}{b_{x_i^{\text{3D}}}} + \frac{\|\delta \hat{y}^{\text{3D}}_{i} - y_{i}^{\text{3D}*} \|_1}{b_{y_i^{\text{3D}}}}  \right. \\
& + & \left. \frac{\| w^{\text{3D}}_{i} - w_{i}^{\text{3D}*} \|_1}{b_{w_i^{\text{3D}}}} + \frac{\| l^{\text{3D}}_{i} - l_{i}^{\text{3D}*} \|_1}{b_{l_i^{\text{3D}}}} + \frac{\| h^{\text{3D}}_{i} - h_{i}^{\text{3D}*} \|_1}{b_{h_i^{\text{3D}}}} \right. \\ 
& + & \left.  \log(b_{x_i^{\text{3D}}}b_{y_i^{\text{3D}}}) + \log(b_{w_i^{\text{3D}}}b_{h_i^{\text{3D}}}b_{l_i^{\text{3D}}})\right) 
\end{aligned}
\end{equation}
Finally, the total loss is formed as 
\begin{equation}
    \mathcal{L}_{\text{total}} = \mathcal{L}_{\text{class}} + \mathcal{L}_{\text{2D}} + \mathcal{L}_{\text{3D}}
\end{equation}

\subsection{Output Decoding and Post-processing}
To decode the network's output, the foreground lidar points are first found by querying the lidar raster using the heatmap output of the class head. Given the set $\mathcal{L}_{\text{f}}$ of lidar points classified as foreground, their corresponding 2D bounding boxes are decoded and processed in a 2D non-maximum suppression step with 0.5 IoU threshold, yielding the reduced set $\mathcal{L}^{\text{2D NMS}}_{\text{f}}$. The 3D bounding boxes of that reduced set are then decoded, and fed into a final bird's eye view NMS step with 0.2 IoU threshold, yielding the final set of output point $\mathcal{L}^{\text{2D/3D NMS}}_{\text{f}}$ and their associated 2D and 3D bounding boxes. An ablation study below shows the relative contribution of the 2D and 3D NMS steps.

\section{Experiments}
\label{sec:experiments}

\begin{table*}[!htbp]
\centering
\caption{Comparison of VRU and vehicle detection performance across different models when running on the ROI defined by the camera FOV, up to 500m in distance.}
\label{tab:detection_performance}
\begin{tabularx}{\textwidth}{@{}l XXX XXXX@{}}

 & \multicolumn{3}{c}{\textbf{VRU bev AP @0.1 IoU}} & \multicolumn{4}{c}{\textbf{Vehicle bev AP @0.1 IoU}} \\
\cmidrule(lr){2-4} \cmidrule(lr){5-8}
\textbf{Model}& \textbf{100-200m} & \textbf{200-300m} & \textbf{300-400m} &  \textbf{100-200m} & \textbf{200-300m} & \textbf{300-400m} & \textbf{400-500m} \\
\midrule
CenterNet & 10.1 & - & -  & 61.1 & 26.5 & 13.2 & 7.6 \\
LaserNet++ & 37.4 & 10.4 & -  & 43.1 & 29.3 & 24.6 & 28.7 \\
Ours (2MP) & \textbf{50.6} & \textbf{34.5} & \textbf{17.5}  & \textbf{71.7} & \textbf{66.8} & \textbf{62.9} & \textbf{55.5} \\ 
\midrule
Ours (8MP) & \textbf{55.5} & \textbf{47.3} & \textbf{29.3}  & \textbf{72.3} & \textbf{72.4} & \textbf{70.7} & \textbf{65.5} \\
\bottomrule
\end{tabularx}
\end{table*}

\begin{table*}[!htbp]
\centering
\caption{Performance evaluation of 2.5D vehicle detection models at various resolutions and distances.}
\label{tab:model_performance}
\label{tab:2.5D metrics comparison}
\begin{tabularx}{\textwidth}{@{}l l XXXX@{}}

 &  & \multicolumn{4}{c}{\textbf{Vehicles 2.5D max f1 @0.5 IoU}} \\
\cmidrule(lr){3-6}
\textbf{Model} & \textbf{Resolution} & \textbf{100-200m} & \textbf{200-300m} & \textbf{300-400m} & \textbf{400-500m} \\
\midrule
CenterNet & 2MP & 62.7 & 56.5 & 51.8 & 46.4 \\
Ours & 2MP & 55.5 & 47.3 & 29.3 & 12.5 \\
\midrule
CenterNet & 8MP & 65.4 & 64.6 & 64.3 & 58.1 \\
Ours & 8MP & \textbf{70.3 }& \textbf{69.5} & \textbf{67.1} & \textbf{61.3} \\
\bottomrule
\end{tabularx}
\end{table*}

\begin{table*}[h]
\centering
\begin{tabularx}{\textwidth}{@{}l l l X X X X X X X X@{}}
 &  &  & \multicolumn{4}{c}{\textbf{VRU bev AP@0.1}} & \multicolumn{4}{c}{\textbf{Vehicle bev AP@0.1}} \\ 
\cmidrule(lr){4-7} \cmidrule(lr){8-11}
\textbf{Resolution}  & \textbf{Supervision} & \textbf{NMS} & \textbf{100-200m} & \textbf{200-300m} & \textbf{300-400m**} & \textbf{400-500m**} & \textbf{100-200m} & \textbf{200-300m} & \textbf{300-400m} & \textbf{400-500m} \\
\midrule
2MP & 3D only & 3D only & 27.9 & 14.4 & 4.2 & 0 & 59.3 & 44.7 & 32.2 & 29.3 \\
2MP & 3D + projected 3D & 3D + 2D & 33.4 & 26.1 & 11.3 & 0 & 58.4 & 52.2 & 51.2 & 47.3 \\
2MP & 3D + 2D & 3D + 2D & 50.6 & 34.5 & 17.5 & 2.6 & 71.7 & 66.8 & 62.9 & 55.5 \\
\midrule
2MP & 3D + 2D & 3D only & 38.1 & 18.8 & 7.2 & 1.3 & 70.4 & 61.3 & 51.4 & 42.3  \\
2MP & 3D + 2D & 2D only & 50.5 & 33.8 & 16.6 & 1.3 & 71.5 & 66.5 & 62.5 & 55.1 \\
2MP & 3D + 2D & 3D + 2D & 50.6 & 34.5 & 17.5 & 2.6 & 71.7 & 66.8 & 62.9 & 55.5 \\
\midrule
8MP*& 3D + 2D & 3D + 2D & \textbf{55.5} & \textbf{47.3} & \textbf{29.3} & \textbf{12.5} & \textbf{72.3} & \textbf{72.4} & \textbf{70.7} & \textbf{65.5} \\
\bottomrule
\end{tabularx}
\smallskip
\textit{*2MP during training, 8MP during inference with rescaled lidar range}
\smallskip
\textit{**sparse VRU labels at these ranges}
\caption{Ablation study for Spotnet. We show the effect of 3D and 2D supervision. In the case of 2D supervision, we show that supervising in image space with true 2D labels considerably improves 3D detection performance, even over using 2D from projected 3D. Using both 2D and 3D nms in postprocessing yields the best results. Finally, a large increase in performance can be achieved by training on 2MP images and testing with 8MP images (with rescaled lidar range). }
\label{tab:results}
\end{table*}

We compare our approach to state of the art approaches that can reasonably scale to ranges of 500m that we are interested in, i.e. range view approaches, both lidar and image-centric. In the absence of publicly available self-driving datasets at these ranges, we use a private Aurora long range dataset.

\definecolor{darked_red}{rgb}{0.55, 0.0, 0.0}

\subsection{Evaluation on Aurora Long Range Dataset}
\label{subsec:aurora long range dataset}
The Aurora long range dataset contains 43,500 five seconds snippets with image, lidar and pose data at 10Hz for training, and 4000 snippets for validation. The image data is from a 30deg FOV long range camera at 8MP resolution. All the lidar data used in these experiments originates from Aurora's FMCW system, with a range of more than 400m. 

We evaluate all methods within the ROI defined by the forward pointing long range camera's 30deg FOV, from 100m to 500m. We use a minimum evaluation range of 100m as nearer distances are for the most part outside of the FOV of the long range camera and lidar. All methods are trained using a single image and 100ms worth of lidar data as input, with the mean point measured time centered on the image timestamp. Our model is trained for 450k iterations, using an Adam optimizer with $8e-4$ starting learning rate and exponential decay of 0.9 every 4000 iterations. During training, the image data is downsampled to 2MP, but is used at either 2MP or 8MP in testing. 

However, when running on 8MP images, the depth values at a given spatial scale as well as the density per pixel area of projected lidar point are no longer consistent with what the model was trained on. To account for this, we apply to following 4 operations: (1) when training on 2MP, we apply a lidar point-wise dropout with a probability of $50\%$; (2) when testing on 8MP images, we remove the point wise dropout thereby keeping the projected point density constant accross image resolution, and (3) rescale the lidar data range values by $0.5$. The resulting depth map will have similar density and range of depth information per unit area at a given scale between 2 and 8MP images. Finally, (4) we undo the range rescaling for each detection before returning them in postprocessing. 

\begin{figure}[H]
  \centering
  \includegraphics[width=\linewidth]{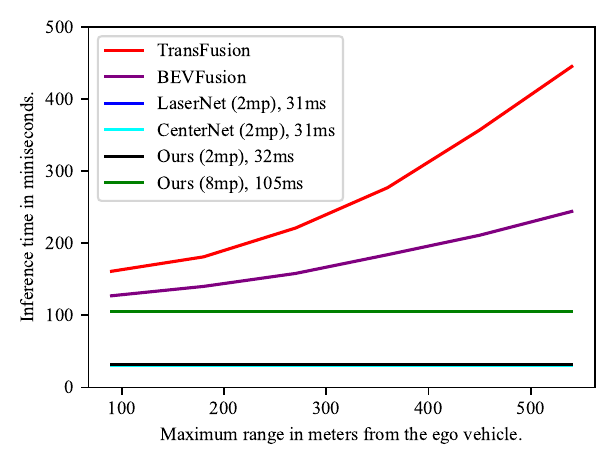}
  \vspace{-10pt}
  \caption{Range based models (SpotNet and CenterNet) enjoy fixed inference times regardless of operating ranges whereas inference times of BEV methods increase with the size of the BEV feature map. The data here was collected on an NVIDIA A10 GPU, where all models were executed with 16-bit floating point precision. Best viewed in color.}
  \label{fig:runtime}
\end{figure}

We begin by comparing inference time between both recent lidar centric BEV and transformer based methods, and various range view (both lidar and image centric) approaches. Figure \ref{fig:runtime} shows that for the long range we are considering here, BEV approaches aren't competitive enough in runtime to be considered real time systems: their runtime scales either $\mathcal{O}(r^2)$ or at best $\mathcal{O}(r)$ with range $r$ (when extending the BEV feature map along only one dimension), while range view methods scale $\mathcal{O}(1)$ with range. Hence in the remainder of the article, we choose to focus solely on range view methods. 

We compare our approach to CenterNet and LaserNet++ in Table \ref{tab:detection_performance}. We evaluate all three method using a 0.1 IoU match in BEV. The loose IoU match requirement we use is motivated by the fact that at long ranges, the presence or absence of object matters much more than their exact positioning or extents. Our method outperform all other methods at all ranges, with the gap widening as the range increases. We observe a significant performance improvement between 2MP and 8MP images at test time.

For the methods that provide 2D bbox for each detection (i.e. our method and CenterNet), we also compute 2.5D metrics, defined as a 0.5 IoU match in 2D bbox and a maximum range error of $10\%$.  Table \ref{tab:2.5D metrics comparison} shows that CenterNet performs better compared to our method on 2.5D metric on 2MP images, given the lesser focus on range accuracy (allowing up to 50m range error at 500m). However, when running on 8MP images (where we applied the same technique detailed above to both models),  CenterNet does not benefit from the increased resolution to the same extent as our approach does and ends up performing worse overall. This is expected as CenterNet regresses the absolute distance to objects directly, and errors will scale with range whereas our method only learns relative deltas and is essentially range invariant. 

\subsection{Ablation Study}

The ablation study detailed in table \ref{tab:results} assesses different configurations of our model, focusing on resolution, supervision modalities, and post-processing.

By far, the component that most impacts the model's performance is the multi-modality supervision: we observe that supervising the model to both regress the 3D box as well as the 2D boxes (from either projected 3D or original 2D labels) considerably increases detection performance over 3D only supervision. We also note that using human labeled 2D boxes significantly improves detection rates over projected 3D. Second, using both the 3D and 2D output in NMS when post-processing detections further improves detection performance. We choose to first run 2D NMS as it is computationally cheaper, followed by 3D NMS on the remaining boxes. We note that the 2D NMS step has the largest impact on performance.
Finally, we observe that training the model on 2MP images while testing on 8MP images significantly improves detection performance across all distances for both VRUs and vehicles.

\subsection{Discussion and Conclusion}

Most of the perception research for self-driving has been focused on the close to medium ($<100m$) range so far. At these ranges, lidar data excels in conveying both accurate geometric and semantic information, as it is dense enough to outline the 3D shape of objects to be detected. This has naturally favored lidar centric methods, as they generally require significantly less data to obtain good 3D detection performance at closer ranges. Here we choose to shift the focus to longer ranges, where the low density of lidar data does not provide any semantic information, and where perception has to strongly rely on imagery. 

Our results suggest that an approach that effectively fuses and leverages each sensor for their respective strength, i.e. accurate range from lidar and rich semantics from images outperforms both lidar centric and image centric baselines. We showed that it is able to leverage image semantic much more effectively than LaserNet++, as well as leverage lidar data much more effectively than CenterNet, and obtain more accurate ranges. We hypothesize that our approach is able to use images data more effectively thanks to a sensor fusion that conserves all image and lidar data, whereas LaserNet++'s approach can only use image data where lidar points are present. At longer ranges with very sparse lidar data, this results in the vast majority of image data being unused. Further, the multi-modal supervision with 2D boxes in image space and 3D boxes in 3D space provides useful gradients to supervise the image features, and results in the model strongly relying on image data.
We believe that our approach is able to leverage lidar data more effectively than CenterNet because it only regresses relative deltas from points to object centers, instead of regressing the absolute distance from camera to the object directly. This renders the method range invariant, and enables it to leverage 8MP imagery effectively at test time, without increasing training costs. 

We believe this approach is efficient and provides reliable 3D long range detection, and is a key step towards a safe autonomous trucking system. Future work will investigate how this method can be used to extend detection beyond lidar range, for example by using lane points in place of lidars point. The method is well suited to achieve this, as it doesn't need to derive any semantics from the 3D points it uses to anchor detections.

\section*{Acknowledgement}

We would like to thank Farshid Moussavi for insightful discussions in particular with the prior work section.

{\small
\bibliographystyle{ieee_fullname}
\bibliography{egbib}
}

\newpage
\appendix
\section*{Appendix}
\begin{figure*}[h]
   \centering
   \includegraphics[width=0.8\textwidth]{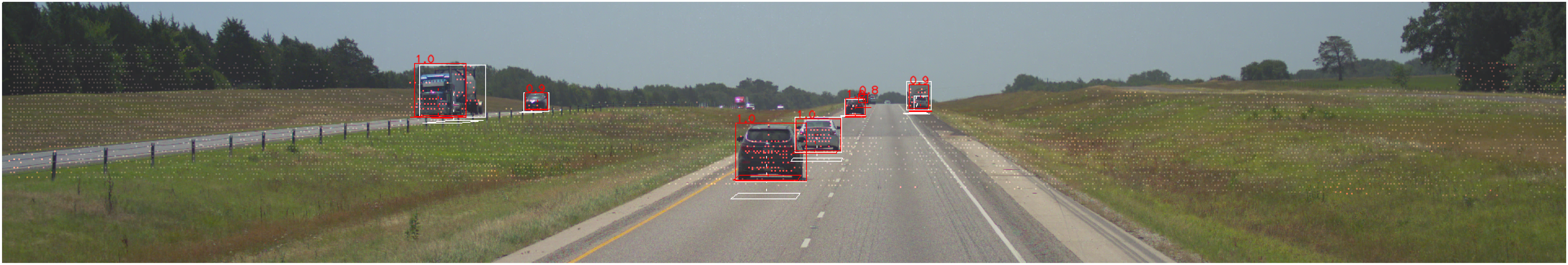}
   \includegraphics[width=0.8\textwidth]{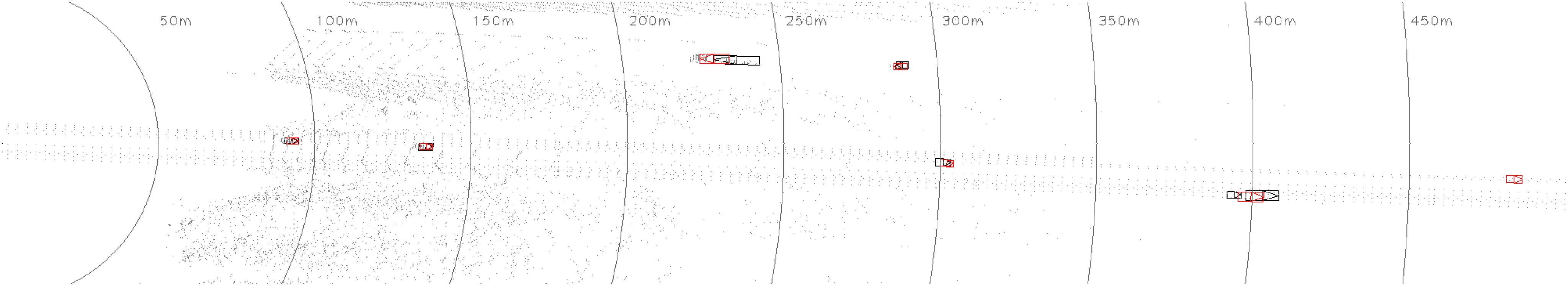}
   \includegraphics[width=0.8\textwidth]{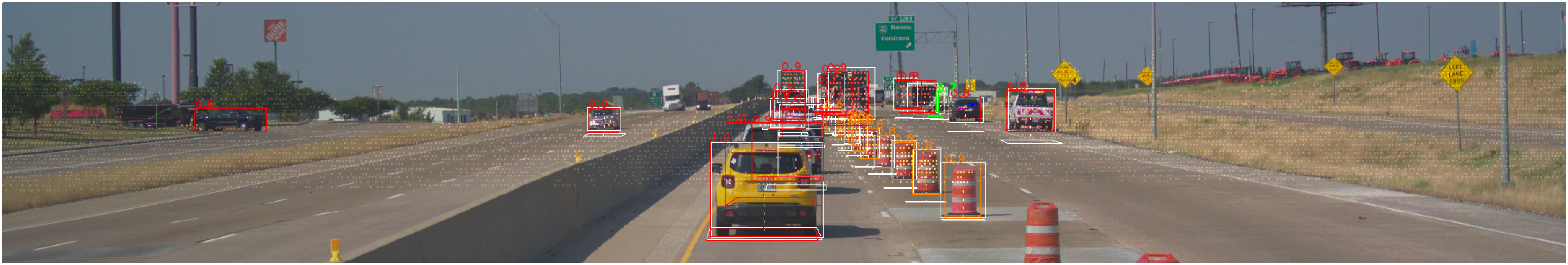}
   \includegraphics[width=0.8\textwidth]{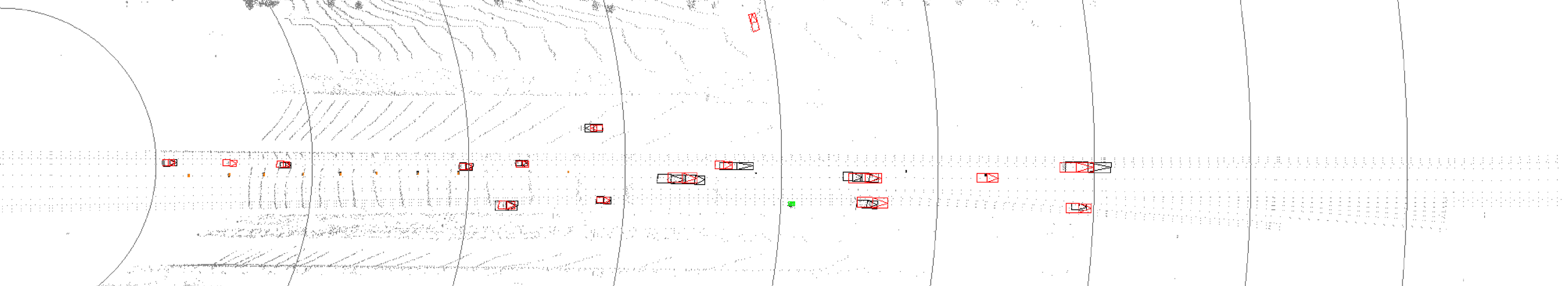}
   \includegraphics[width=0.8\textwidth]{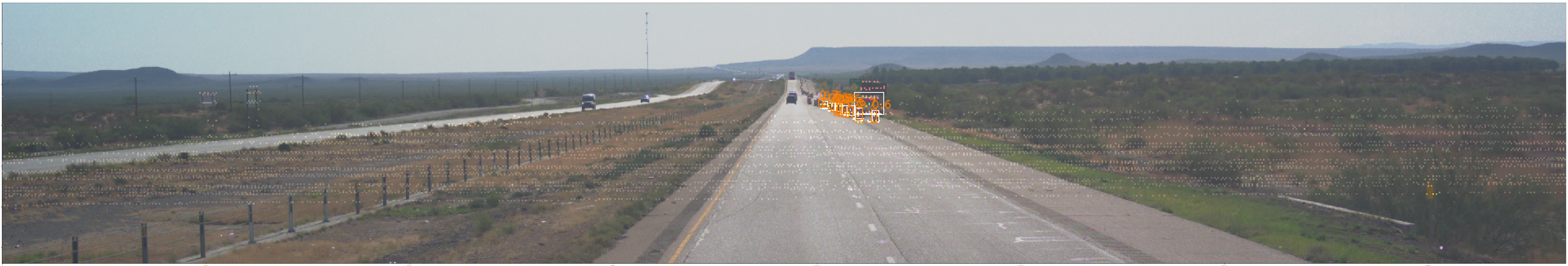}
   \includegraphics[width=0.8\textwidth]{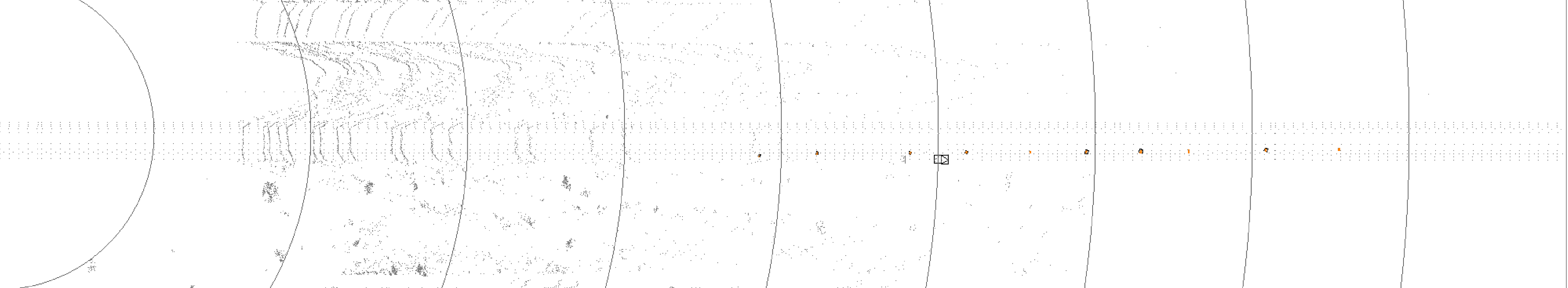}

    \caption{Examples detections on our validation dataset, out to ranges of 450m, anchored on FMCW lidar points. The furthest detections only have 1-2 lidar points. Colors of the bounding boxes of detections: 
    {\color{red} vehicle}, {\color{green} pedestrian}, {\color{orange} construction}; all the linked 2d/3d labels are displayed in white. }
    \label{fig: Detection examples}
\end{figure*}

We include additional visualizations of spotnet detection outputs at long ranges here in \cref{fig: Detection examples}.

\end{document}